\documentclass[runningheads]{llncs}
\usepackage{graphicx}

\usepackage{tikz}
\usepackage{comment}
\usepackage{amsmath,amssymb} 
\usepackage{color}
\usepackage{float}
\usepackage{wrapfig}
\usepackage{booktabs}
\floatstyle{boxed} 

\usepackage[accsupp]{axessibility}  
\usepackage[pagebackref,breaklinks,colorlinks]{hyperref}

\usepackage[width=122mm,left=12mm,paperwidth=146mm,height=193mm,top=12mm,paperheight=217mm]{geometry}
\usepackage{mmstyle}

\begin{document}
\pagestyle{headings}
\mainmatter
\def\ECCVSubNumber{3999}  

\title{Monocular 3D Object Reconstruction \\ with GAN Inversion} 

\titlerunning{MeshInversion}
\author{Junzhe Zhang\inst{1,3} \and
Daxuan Ren\inst{1,3} \and 
Zhongang Cai\inst{1,3} \and \\ 
Chai Kiat Yeo\inst{2} \and 
Bo Dai\inst{4}\thanks{Bo Dai completed this work when he was with S-Lab, NTU.} \and
Chen Change Loy\inst{1}}
\authorrunning{Zhang et al.}
\institute{S-Lab, Nanyang Technological University  \and Nanyang Technological University  \and SenseTime Research \and Shanghai AI Laboratory \\
\email{\{junzhe001,daxuan001,caiz0023\}@e.ntu.edu.sg,\\ \{asckyeo,ccloy\}@ntu.edu.sg, \{daibo\}@pjlab.org.cn}}
\maketitle

\begin{abstract}

Recovering a textured 3D mesh from a monocular image is highly challenging, particularly for in-the-wild objects that lack 3D ground truths.
In this work, we present \textbf{MeshInversion}, a novel framework to improve the reconstruction by exploiting the \textit{generative prior} of a 3D GAN pre-trained for 3D textured mesh synthesis.
Reconstruction is achieved by searching for a latent space in the 3D GAN that best resembles the target mesh in accordance with the single view observation.
Since the pre-trained GAN encapsulates rich 3D semantics in terms of mesh geometry and texture,
searching within the GAN manifold thus naturally regularizes the realness and fidelity of the reconstruction.
Importantly, such regularization is directly applied in the 3D space, providing crucial guidance of mesh parts that are unobserved in the 2D space.
Experiments on standard benchmarks show that our framework obtains faithful 3D reconstructions with consistent geometry and texture across both observed and unobserved parts.
Moreover, it generalizes well to meshes that are less commonly seen, such as the extended articulation of deformable objects. Code is released at \textcolor{magenta}{\url{https://github.com/junzhezhang/mesh-inversion}}.

\end{abstract}

\section{Introduction}
\label{sec:intro}


\vspace{-0.2cm}
\begin{figure}[ht]
	\centering
	\includegraphics[width=0.95\linewidth]{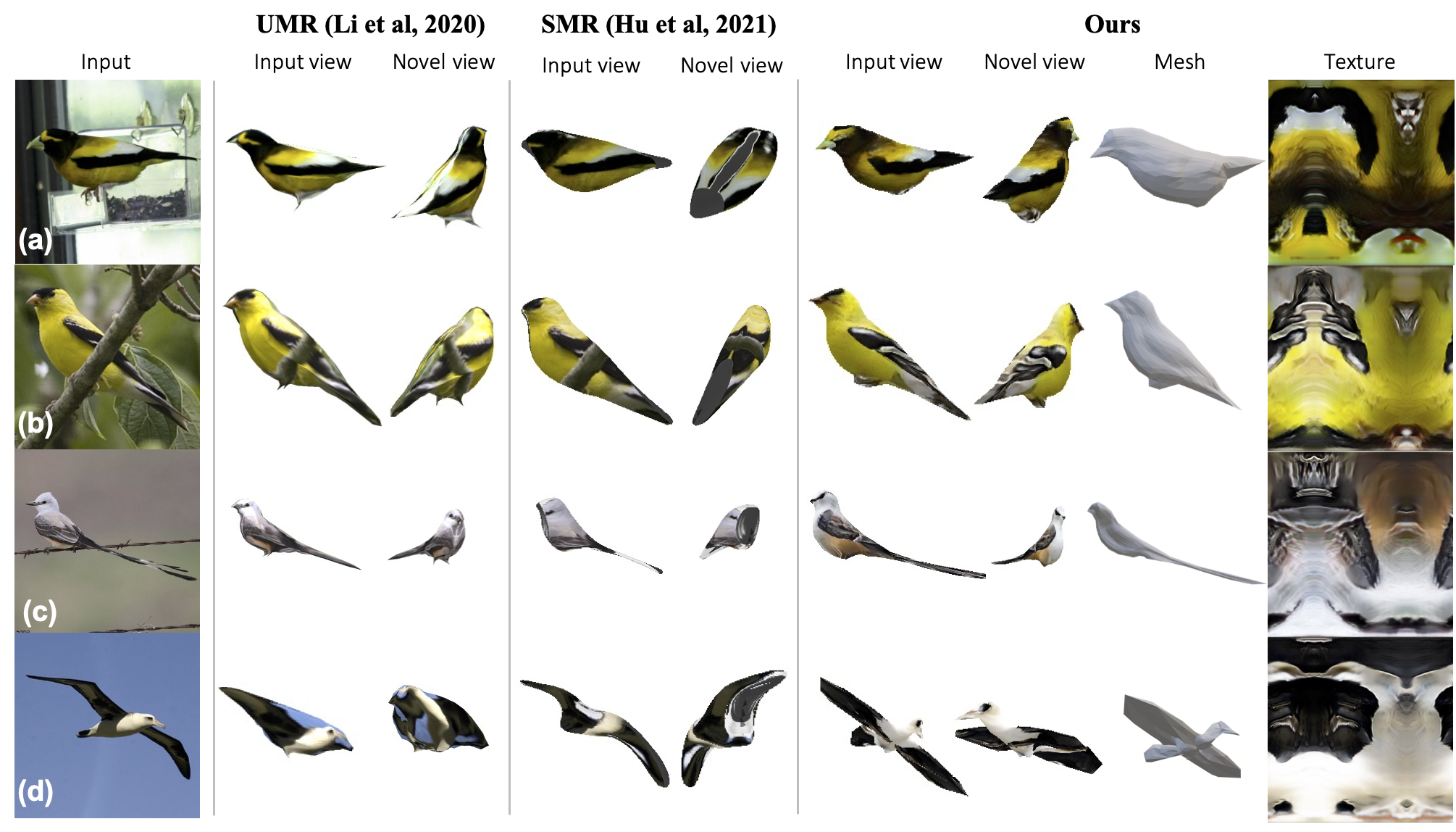}
	\caption{We propose an alternative approach to monocular 3D reconstruction by exploiting generative prior encapsulated in a pre-trained GAN. Our method has three major advantages: \textbf{1)} It reconstructs highly faithful and realistic 3D objects, even when observed from novel views; \textbf{2)} The reconstruction is robust against occlusion in (b);
	\textbf{3)} The method generalizes reasonably well to less common shapes, such as birds with (c) extended tails or (d) open wings. 
	}
	\label{fig:teaser}
\end{figure}

We consider the task of recovering the 3D shape and texture of an object from its monocular observation.
A key challenge in this task is the lack of 3D or multi-view supervision due to the prohibitive cost of data collection and annotation for object instances in the wild.

Prior attempts resort to weak supervision based on 2D silhouette annotations of monocular images to solve this task. For instance, Kanazawa~\etal.~\cite{kanazawa2018learning} propose the use of more readily available 2D supervisions including keypoints as the supervision.
To further relax the supervision constraint, several follow-up studies propose to learn the 3D manifold in a self-supervised manner, only requiring single-view images and their corresponding masks for training~\cite{li2020self,goel2020shape,bhattad2021view,hu2021self}. 
Minimizing the reconstruction error in the 2D domain tends to ignore the overall 3D geometry and back-side appearance, leading to a shortcut solution that may look plausible only from the input viewpoint, \eg, SMR~\cite{hu2021self} in Fig.~\ref{fig:teaser}  (a)(c).
While these methods compensate the relaxed supervision by exploiting various forms of prior information, 
\eg, categorical semantic invariance~\cite{li2020self} and interpolated consistency of the predicted 3D attributes~\cite{hu2021self}, this task remains challenging. 

In this work, we propose a new approach, \textbf{MeshInversion}, that is built upon generative prior possessed by Generative Adversarial Networks (GANs)~\cite{goodfellow2014generative}. GANs are typically known for their exceptional ability to capture comprehensive knowledge~\cite{brock2018large,karras2019style,shu20193d}, empowering the success of GAN inversion in image restoration \cite{gu2020image,pan2021exploiting} and point cloud completion~\cite{zhang2021unsupervised}.
We believe that by training a GAN to synthesize 3D shapes in the form of a topology-aligned texture and deformation map, one could enable the generator to capture rich prior knowledge of a certain object category, including high-level semantics, object geometries, and texture details.

We propose to exploit the appealing generative prior through GAN inversion. Specifically, our framework finds the latent code of the pre-trained 3D GAN that best recovers the 3D object in accordance with the single-view observation. Given the RGB image and its associated silhouette mask  estimated by an off-the-shelf segmentation model, the latent code is optimized towards minimizing 2D reconstruction losses by rendering the 3D object onto the 2D image plane.
Hence, the latent manifold of the 3D GAN \textit{implicitly} constrains the reconstructed 3D shape within the realistic boundaries, whereas minimization of 2D losses \textit{explicitly} drives the 3D shape towards a faithful reflection of the input image.  

Searching for the optimal latent code in the GAN manifold for single-view 3D object reconstruction is non-trivial due to following challenges: 1) Accurate camera poses are not always available for real-world applications. Inaccurate camera poses easily lead to reprojection misalignment and thus erroneous reconstruction. 2) Existing geometric losses that are computed between 2D masks inevitably discretize mesh vertices into a grid of pixels during rasterization. Such discretization typically makes the losses less sensitive in reflecting the subtle geometric variations in the 3D space.
To address the misalignment issue, we propose a \textbf{Chamfer Texture Loss}, which relaxes the one-to-one pixel correspondences in existing losses and allows the match to be found within a local region. By jointly considering the appearance and positions of image pixels or feature vectors, it provides a robust texture distance despite inaccurate camera poses and in the presence of high-frequency textures.
To improve the geometric sensitivity, we propose a \textbf{Chamfer Mask Loss}, which intercepts the rasterization process and computes the Chamfer distance between the projected vertices before discretization to retain information, with the foreground pixels of the input image projected to the same continuous space. Hence, it is more sensitive to small variations in shape and offers a more accurate gradient for geometric learning.

MeshInversion demonstrates compelling performance for 3D reconstruction from real-world monocular images.
Even with the assumption of inaccurate masks and camera poses, our method still gives highly plausible and faithful 3D reconstruction in terms of both appearance and 3D shape, as depicted in Fig.~\ref{fig:teaser}.
It achieves state-of-the-art results on the perceptual metric, \ie, FID, when evaluating the textured mesh from various viewpoints, and is on-par with the existing CMR-based frameworks in terms of geometric accuracy. 
In addition, while its holistic understanding of the objects benefits from the generative prior, 
it not only gives a realistic recovery of the back-side texture but also generalizes well in the presence of occlusion, \eg, Fig.~\ref{fig:teaser} (b).
Furthermore, MeshInversion also demonstrates significantly better generalization for 3D shapes that are less commonly seen, such as birds with open wings and long tails, as shown in Fig.~\ref{fig:teaser} (d) and (c) respectively. 
\section{Related Work}
\label{sec:related_work}

\noindent
\textbf{Single-view 3D Reconstruction.} 
Many methods have been proposed to recover the 3D information of an object, such as its shape and texture, from a single-view observation. Some methods use image-3D object pairs~\cite{wang2018pixel2mesh,pan2019deep,mescheder2019occupancy,rematas2021sharf} or multi-view images~\cite{niemeyer2020differentiable,liu2019soft,yariv2020multiview,wang2021neus,oechsle2021unisurf} for training, which limit the scenarios to synthetic data. Another line of work fits the parameters of a 3D prior morphable model, \eg, SMPL for humans and 3DMM for faces~\cite{gecer2019ganfit,sanyal2019learning,kanazawa2018end}, which are typically expensive to build and difficult to extend to various natural object categories. 

To relax the constraints on supervision, CMR~\cite{kanazawa2018learning} reconstructs category-specific textured mesh by training with a collection of monocular images and associated 2D supervisions, \ie, 2D key-points, camera poses, and silhouette masks.
Thereafter, several follow-up studies further relax the supervision, \eg, masks only, and improves the reconstruction results by exploiting different forms of prior.
Specifically, they incorporate the prior by enforcing various types of cycle consistencies, such as texture cycle consistency~\cite{li2020self,bhattad2021view}, rotation adversarial cycle consistency~\cite{bhattad2021view}, and interpolated consistency~\cite{hu2021self}. Some of these methods also leverage external information, \eg, category-level mesh templates~\cite{goel2020shape,bhattad2021view}, and semantic parts provided by an external SCOPS model~\cite{li2020self}.
In parallel, Shelf-Sup~\cite{ye2021shelf} first gives a coarse volumetric prediction, and then converts the coarse volume into a mesh followed by test-time optimization. Without categorical mesh templates in existing approaches, this design demonstrates its scalability to categories with high-genus meshes, \eg, chairs and backpacks.

For texture modeling, direct regression of pixel values in the UV texture map often leads to blurry images~\cite{goel2020shape}. Therefore, the mainstream approach is to regress pixel coordinates, \ie, learning \textit{texture flow} from the input image to the texture map. Although texture flow is easier to regress and usually provides a vivid front view result, it often fails to generalize well to novel views or occluded regions.
Our approach directly predicts the texture pixel values by incorporating a pre-trained GAN. In contrast to the texture flow approach, it benefits from a holistic understanding of the objects given the generative prior and offers high plausibility and fidelity at the same time. 

\noindent
\textbf{GAN Inversion.} A well-trained GAN usually captures useful statistics and semantics underlying the training data. In the 2D domain, GAN prior has been explored extensively in various image restoration and editing tasks \cite{bau2019seeing,gu2020image,pan2021exploiting}. GAN inversion, the common method in this line of work, finds a latent code that best reconstructs the given image using the pre-trained generator.
Typically, the target latent code can be obtained via gradient descent~\cite{ma2018invertibility,lipton2017precise}, projected by an additive encoder that learns the inverse mapping of a GAN \cite{bau2020semantic}, or a combination of them~\cite{zhu2020domain}.
There are recent attempts to apply GAN inversion in the 3D domain. Zhang~\etal.~\cite{zhang2021unsupervised} use a pre-trained point cloud GAN to address shape completion in the canonical pose, giving remarkable generalization for out-of-domain data such as real-world partial scans. Pan~\etal.~\cite{pan20202d} recover the geometric cues from pre-trained 2D GANs and achieve exceptional reconstruction results, but the reconstructed shapes are limited to 2.5D due to limited poses that 2D GANs can synthesize. In this work, we directly exploit the prior from a 3D GAN to reconstruct the shape and texture of complete 3D objects.

\noindent
\textbf{Textured Mesh Generation.} 3D object generation approaches that use voxels~\cite{wu2016learning,girdhar2016learning,smith2017improved,zhu2018visual,xie2018learning} or point clouds~\cite{achlioptas2018learning,shu20193d} typically require some form of 3D supervision and are unfriendly for modeling texture. 
Chen~\etal.~\cite{chen2019learning} propose DIB-R, a GAN framework for textured mesh generation, where 3D meshes are differentiably rendered into 2D images and discriminated with multi-view images of synthetic objects. Later on, Henderson~\etal.~\cite{henderson2020leveraging} relax the multi-view restriction and propose a VAE framework~\cite{kingma2013auto} that leverages a collection of single-view natural images. The appearance is parameterized by face colors instead of texture maps, limiting the visual detail of generated objects.
Under the same setting, ConvMesh~\cite{pavllo2020convolutional} achieves more realistic 3D generations by generating 3D objects in the form of topology-aligned texture maps and deformation maps in the UV space, where discrimination directly takes place in the UV space against pseudo ground truths. Pseudo deformation maps are obtained by overfitting a mesh reconstruction baseline on the training set. Subsequently, the associated pseudo texture maps can then be obtained by projecting natural images on the UV space. Our proposed method is built upon a pre-trained ConvMesh model to incorporate its generative prior in 3D reconstruction.
\section{Approach}
\label{sec:approach}

\begin{figure}[t]
	\centering
	\includegraphics[width=0.93\linewidth]{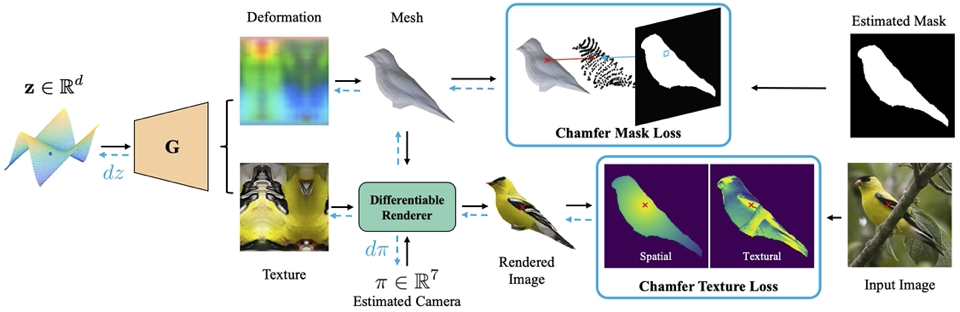}
	\caption{ \textbf{Overview of the MeshInversion framework.}
	We reconstruct a 3D object from its monocular observation by incorporating a pre-trained textured 3D GAN $\mathbf{G}$.
	We search for the latent code $\mathbf{z}$ and fine-tune the imperfect camera $\mathbf{\pi}$ that minimizes 2D reconstruction losses via gradient descent.
    To address the intrinsic challenges associated with 3D-to-2D degradation, we propose two Chamfer-based losses:
	1) \textbf{Chamfer Texture Loss} (Sec~\ref{sec:fc_loss}) relaxes the pixel-wise correspondences between two RGB images or feature maps, and factorizes the pairwise distance into spatial and textural distance terms. 
	We illustrate the distance maps between one anchor point from the rendered image to the input image, where brighter regions correspond to smaller distances.
	2) \textbf{Chamfer Mask Loss} (Sec~\ref{sec:cm_loss}) intercepts the discretization process and computes the Chamfer distance between the projected vertices and the foreground pixels projected to the same continuous space. No labeled data is assumed during inference: the mask and camera are estimated by off-the-shelf pre-trained models.}
	\label{fig:framework}
\end{figure}

\noindent
\textbf{Preliminaries.} 
We represent a 3D object as a textured triangle mesh $\mathbf{O} \equiv (\mathbf{V}, \mathbf{F}, \mathbf{T})$, where $\mathbf{V} \in \Rbb^{|\mathbf{v}| \times 3}$ represents the location of the vertices, $\mathbf{F}$ represents the faces that define the fixed connectivity of vertices in the mesh, and $\mathbf{T}$ represents the texture map.
An individual mesh is iso-morphic to a 2-pole sphere, and thus we model the deformation $\Delta \mathbf{V}$ from the initial sphere template, and then obtain the final vertex positions by $\mathbf{V} = \mathbf{V}_{sphere} + \Delta \mathbf{V}$. Previous methods~\cite{kanazawa2018learning,li2020self,goel2020shape} mostly regress deformation of individual vertices via a fully connected network (MLP). In contrast, recent studies have found that using a 2D convolutional neural network (CNN) to learn a deformation map in the UV space would benefit from consistent semantics across the entire category~\cite{pavllo2020convolutional,bhattad2021view}. 
In addition, the deformation map $\mathbf{S}$ and the texture map $\mathbf{T}$ are topologically aligned, so both the values can be mapped to the mesh via the same predefined mapping function.

We assume a weak-perspective camera projection, where the camera pose $\mathbf{\pi}$ is parameterized by scale $\mathbf{s} \in \Rbb$, translation $\mathbf{t} \in \Rbb^2$, and rotation in the form of quaternion $\mathbf{r} \in \Rbb^4$. We use DIB-R~\cite{chen2019learning} as our differentiable renderer. We denote $\mathbf{I} = R(\mathbf{S}, \mathbf{T}, \mathbf{\pi})$ as the image rendering process.
Similar to previous baselines~\cite{kanazawa2018learning,li2020self,hu2021self,goel2020shape}, we enforce reflectional symmetry along the $x$ axis, which both benefits geometric performance and reduces computation cost.

\subsection{Reconstruction with Generative Prior}
\label{sec:mesh_inversion}
Our study presents the first attempt to explore the effectiveness of generative prior in monocular 3D construction. Our framework assumes a pre-trained textured 3D GAN. In this study, we adopt ConvMesh~\cite{pavllo2020convolutional}, which is purely trained with 2D supervisions from single-view natural images. With the help of GAN prior, our goal is to recover the geometry and appearance of a 3D object from a monocular image and its associated mask. Unlike SMR~\cite{hu2021self} that uses ground truth masks, our method takes silhouettes estimated by an off-the-shelf segmentation model~\cite{kirillov2020pointrend}.

Next, we will detail the proposed approach to harness the meaningful prior, such as high-level semantics, object geometries, and texture details, from this pre-trained GAN to achieve plausible and faithful recovery of 3D shape and appearance. Note that our method is not limited to ConvMesh, and other pre-trained GANs that generate textured meshes are also applicable.
More details of ConvMesh can be found in the supplementary materials.

\noindent
\textbf{Pre-training Stage.} Prior to GAN inversion, we first pre-train the textured GAN on the training split to capture desirable prior knowledge for 3D reconstruction.
As discussed in Sec.~\ref{sec:related_work}, the adversarial training of ConvMesh takes place in the UV space, where generated deformation maps and texture maps are discriminated against their corresponding pseudo ground truth.
In addition to the UV space discrimination, we further enhance the photorealism of the generated 3D objects by introducing a discriminator in the image space, following the architecture of PatchGAN as in~\cite{isola2017image}. 
The loss functions for the pre-training stage are shown as follows, where $D_{uv}$ and  $D_I$ refer to the discriminators in the UV space and image space respectively, and $\lambda_{uv}$ and $\lambda_{I}$ are the corresponding weights. We use least-squares losses following~\cite{mao2017least}. An ablation study on image space discrimination can be found in the supplementary materials.
\begin{equation}
   \mathcal{L}_G = \lambda_{uv}\Ebb_{\mathbf{z} \sim P_{\mathbf{z}}} [ (D_{uv}(G(\mathbf{z}))-1)^2 ] + \lambda_{I}\Ebb_{\mathbf{z} \sim P_\mathbf{z}(\mathbf{z})} [ (D_{I}(R(G(\mathbf{z})),\mathbf{\pi})-1)^2 ].
\end{equation}
\begin{equation}
   \mathcal{L}_{D_{uv}} = \Ebb_{\mathbf{S}, \mathbf{T} \sim P_{pseudo}} [ (D_{uv}(\mathbf{S}, \mathbf{T})-1)^2 ] + \Ebb_{\mathbf{z} \sim P_\mathbf{z}(\mathbf{z})} [ (D_{uv}(G(\mathbf{z})))^2 ].
\end{equation}
\begin{equation}
   \mathcal{L}_{D_{I}} = \Ebb_{\mathbf{I} \sim P_{data}} [ (D_{I}(\mathbf{I})-1)^2 ] + \Ebb_{\mathbf{z} \sim P_\mathbf{z}(\mathbf{z})} [ (D_{I}(R(G(\mathbf{z})),\mathbf{\pi})^2 ].
\end{equation}

\noindent
\textbf{Inversion Stage.} We now formally introduce GAN inversion for single-view 3D reconstruction.
Given a pre-trained ConvMesh that generates a textured mesh from a latent code, $\mathbf{S}, \mathbf{T}  = G(\mathbf{z})$, we aim to find the $\mathbf{z}$ that best recovers the 3D object from the input image $\mathbf{I}_{in}$ and its silhouette mask $\mathbf{M}_{in}$.
Specifically, we search for such $\mathbf{z}$ via gradient descent towards minimizing the overall reconstruction loss $\mathcal{L}_{inv}$, which can be denoted by
\begin{equation}
\label{eq:inv_z}
    \mathbf{z}^* = \argmin_{\mathbf{z}} \mathcal{L}_{inv}(R(G(\mathbf{z}),\mathbf{\pi}),\mathbf{I}_{in}).
\end{equation}

Given the single-view image and the associated mask, we would need to project the reconstructed 3D object to the observation space for computing $\mathcal{L}_{inv}$. However, such 3D-to-2D degradation is non-trivial.
Unlike existing image-based GAN inversion tasks where we can always assume pixel-wise image correspondence in the observation space, rendering 3D objects in the canonical frame onto the image space is explicitly controlled via camera poses. For real-world applications, unfortunately, perfect camera poses are not always available to guarantee such pixel-wise image correspondence. 
While concurrently optimizing the latent code and the camera pose from scratch seems a plausible approach, this often suffers from camera-shape ambiguity~\cite{li2020self} and leads to erroneous reconstruction. To this end, we initialize the camera with a camera pose estimator (CMR~\cite{kanazawa2018learning}), which can be potentially inaccurate, and jointly optimize it in the course of GAN inversion, for which we have:
\begin{equation}
\label{eq:inv_zpi}
    \mathbf{z}^*,\mathbf{\pi}^* = \argmin_{\mathbf{z},\mathbf{\pi}} \mathcal{L}_{inv}(R(G(\mathbf{z}),\mathbf{\pi}),\mathbf{I}_{in}). 
\end{equation}

As the 3D object and cameras are constantly optimized throughout the inversion stage, it is infeasible to assume perfect image alignment.
In addition, the presence of high-frequency textures, \eg, complex bird feathers, often leads to blurry appearance even with slight discrepancies in pose. Consequently, it calls for a robust form of texture loss in Sec~\ref{sec:fc_loss}.

\subsection{Chamfer Texture Loss}
\label{sec:fc_loss}
To facilitate searching in the GAN manifold without worrying about blurry reconstructions, we reconsider the appearance loss by relaxing the pixel-aligned assumption in existing low-level losses.
Taking inspiration from the point cloud data structure, we treat a 2D image as a set of 2D colored points, which have both appearance attributes, \ie, RGB values, and spatial attributes, the values of which relate to their coordinates in the image grid.
Thereafter, we aim to measure the dissimilarity between the two colored point sets via Chamfer distance,
\begin{equation}
    \mathcal{L}_{CD}(\Sbb_1, \Sbb_2) = \frac{1}{|\Sbb_1|}\sum_{x\in \Sbb_1}\min_{y\in \Sbb_2}\mathbf{D}_{xy} + \frac{1}{|\Sbb_2|}\sum_{y \in \Sbb_2}\min_{x\in \Sbb_1}\mathbf{D}_{yx}.
    \label{eq:cd}
\end{equation}
Intuitively, defining the pairwise distance between pixel $x$ and pixel $y$ in the two respective images should jointly consider their appearance and location.
In this regard, we factorize the overall pairwise distance $\mathbf{D}_{xy}$ into an appearance term $\mathbf{D}_{xy}^a$ and a spatial term $\mathbf{D}_{xy}^s$, both of which are L2 distance. Like conventional Chamfer distance, single-sided pixel correspondences are determined by column-wise or row-wise minimum in the distance matrix $\mathbf{D}$.

Importantly, we desire the loss to be tolerant and only tolerant of local misalignment, as large misalignment will potentially introduce noisy pixel correspondences that may jeopardize appearance learning.
Inspired by the focal loss for detection~\cite{lin2017focal}, we introduce an exponential operation in the spatial term to penalize those spatially distant pixel pairs. 
Therefore, we define the overall distance matrix $\mathbf{D}  \in \Rbb ^ {|\Sbb_1| \times |\Sbb_2|}$ as follows 
\begin{equation}
    \mathbf{D} = \max((\mathbf{D}^s+\epsilon_s)^{\alpha},1) \otimes (\mathbf{D}^a + \epsilon_a),
    \label{eq:focal_d}
\end{equation}
where $\mathbf{D}^a$ and $\mathbf{D}^s$ are the appearance distance matrix and spatial distance matrix respectively; $\otimes$ denotes element-wise product; $\epsilon_s$ and $\epsilon_a$ are residual terms to avoid incorrect matches with identical location or identical pixel value respectively; $\alpha$ is the scaling factor for flexibility.
Specifically, we let $\epsilon_s < 1$ so that the spatial term remains one when two pixels are slightly misaligned. Note that the spatial term is not differentiable and it only serves as a weight matrix for appearance learning.
By substituting the resulting $\mathbf{D}$ into Eq.~\ref{eq:cd}, we thus have the final formulation of our proposed \textbf{Chamfer Texture Loss}, denoted as $\mathcal{L}_{CT}$.

The proposed relaxed formulation provides a robust measure of texture distance, which effectively eases searching of the target latent code while preventing blurry reconstructions; in return, 
although $\mathcal{L}_{CT}$ only concerns about local patch statistics but not photorealism, the use of GAN prior is sufficient to give realistic predictions. Besides, the GAN prior also allows computing $\mathcal{L}_{CT}$ with a down-sampled size of colored points. In practice, we randomly select 8096 pixels from each image as a point set.

The proposed formulation additionally gives flexible control between appearance and spatial attributes. The appearance term is readily extendable to accept misaligned feature maps to achieve more semantically faithful 3D reconstruction.
Specifically, we apply the Chamfer texture loss between the (foreground) feature maps extracted with a pre-trained VGG-19 network~\cite{simonyan2014very} from the rendered image and the input image. It is worth noting that the feature-level Chamfer texture loss is somewhat related to the contextual loss~\cite{mechrez2018contextual}, which addresses the misalignment issue for image transfer. The key difference is that the contextual loss only considers the feature distances but ignores their locations. We compare against the contextual loss in the experiment.

\subsection{Chamfer Mask Loss}
\label{sec:cm_loss}
Conventionally, the geometric distance is usually computed between two binary masks in terms of L1 or IoU loss~\cite{kanazawa2018learning,goel2020shape,li2020self,hu2021self}. 
However, obtaining the mask of the reconstructed mesh usually involves rasterization that discretizes the mesh into a grid of pixels. This operation inevitably introduces information loss and thus inaccurate supervision signals.
This is particularly harmful to a well-trained ConvMesh, the shape manifold of which is typically smooth.
Specifically, a small perturbation in $\mathbf{z}$ usually corresponds to a slight variation in the 3D shape, which may translate to an unchanged binary mask.
This usually leads to an insensitive gradient for back-propagation, which undermines geometric learning. We analyze the sensitivity of existing losses in the experiment.

To this end, we propose a \textbf{Chamfer Mask Loss}, or $\mathcal{L}_{CM}$, to compute the geometric distance in an unquantized 2D space.
Instead of rendering the mesh into a binary mask, we directly project the 3D vertices of the mesh onto the image plane, $\Sbb_v = P(\mathbf{S},\mathbf{T},\mathbf{\pi})$. For the foreground mask, we obtain the positions of the foreground pixels by normalizing their pixel coordinates in the range of $[-1,1]$, denoted as $\Sbb_f$. Thereafter, we compute the Chamfer distance between $\Sbb_v$ and $\Sbb_f$ as the Chamfer mask loss.
Note that one does not need to distinguish visible and occluded vertices, as eventually they all fall within the rendered silhouette. The bidirectional Chamfer distance between the sparse set $\Sbb_v$ and dense set $\Sbb_f$ would regularize the vertices from highly uneven deformation.

\subsection{Overall Objective Function} 
We apply the pixel-level Chamfer texture loss $\mathcal{L}_{pCT}$ and the feature-level one $\mathcal{L}_{fCT}$ as our appearance losses, and the Chamfer mask loss $\mathcal{L}_{CM}$ as our geometric loss. Besides, we also introduce two regularizers:
the smooth loss $\mathcal{L}_{smooth}$ that encourages neighboring faces to have similar normals, \ie, low cosine; 
the latent space loss $\mathcal{L}_{z}$ that regularizes the L2 norm of $\mathbf{z}$ to ensure Gaussian distribution. In summary, the overall objective function is shown in Eq.~\ref{eq:overall_loss}.
\begin{equation}
\begin{aligned}  
    \mathcal{L}_{inv} = \mathcal{L}_{pCT} +
    \mathcal{L}_{fCT} + \mathcal{L}_{CM}
    + \mathcal{L}_{smooth} + \mathcal{L}_{z}.
    \label{eq:overall_loss}
\end{aligned}  
\end{equation}
\section{Experiments}
\label{sec:exp}

\noindent
\textbf{Datasets and Experimental Setting.} 
We primarily evaluate MeshInversion on CUB-200-2011 dataset~\cite{wah2011caltech}. It consists of 200 species of birds with a wide range of shapes and feathers, making it an ideal benchmark to evaluate 3D reconstruction in terms of both geometric and texture fidelity.
Apart from the organic shapes like birds, we also validate our method on 11 man-made rigid car categories from PASCAL3D+~\cite{xiang2014beyond}.

We use the same train-validation-test split as provided by CMR~\cite{kanazawa2018learning}. The images in both datasets are annotated with foreground masks and camera poses. Specifically, we pre-train ConvMesh on the pseudo ground truths derived from the training split following a class conditional setting~\cite{pavllo2020convolutional}.
During inference, we conduct GAN inversion on the test split without assuming additional labeled data compared to existing methods. We use the silhouette masks predicted by an off-the-shelf instance segmentation method PointRend~\cite{kirillov2020pointrend} pre-trained on COCO~\cite{lin2014microsoft}, which gives an IoU of 0.886 against ground truth masks. We use camera poses predicted by CMR~\cite{kanazawa2018learning}, which can be inaccurate. In particular, the poses estimated yields 6.03 degree of azimuth error and 4.33 degree of elevation error compared to the ground truth cameras via structure-from-motion (SfM).
During evaluation, we report quantitative results based on ground truth masks.

\noindent
\textbf{Evaluation Strategy.} Since there are no 3D ground truths available for CUB, we evaluate MeshInversion against various baselines from three aspects: 1) We evaluate the geometry accuracy in the 2D domain by IoU between the rendered masks and the ground truths. 
2) We evaluate the appearance quality by the image synthesis metric FID~\cite{heusel2017gans}, which compares the distribution of test set images and the render of reconstruction. 
Since a plausible 3D shape should look photo-realistic observed from multiple viewpoints, we report both single-view FID (\textbf{FID$_{1}$}) and multi-view FIDs. Following SMR~\cite{hu2021self} and View-gen~\cite{bhattad2021view}, we render our reconstructed 3D shape from 12 different views (\textbf{FID$_{12}$}), which covers azimuth from 0$^{\circ}$ to 360$^{\circ}$ at an interval of 30$^{\circ}$. We additionally report \textbf{FID$_{10}$} since the exact front view (90$^{\circ}$) and the exact back view (270$^{\circ}$) are rarely seen in CUB. Note that this is in favour of existing methods that do not use any GAN prior as ours.
3) Apart from extensive qualitative results, we conduct a user study to evaluate human preferences in terms of both shape and appearance. 
For PASCAL3D+, it provides approximated 3D shapes using a set of 10 CAD models, which allows us to evaluate geometric performance in terms of 3D IoU.

\begin{table}[t]
\setlength{\tabcolsep}{4.5pt}
\centering
\caption{Quantitative results on CUB show the effectiveness of applying generative prior in 3D reconstruction. As all the baseline methods are regression-based whereas our method involves optimization during inference, we report both baselines and test-time optimization (TTO) results for existing methods, if applicable, with access to masks estimated by PointRend~\cite{kirillov2020pointrend}. SMR baseline uses ground truth mask, and it shows noticeable IoU drop with estimated mask.
$^{\dagger}$: We report results from \cite{bhattad2021view} since no implementation released; the results are based on ground truth cameras, whereas our method optimizes from imperfect cameras.
}
\small
\begin{tabular}{lcc|cccc}
\hline
Methods & TTO & input mask & IoU $\uparrow$ & FID$_{1}$ $\downarrow$  & FID$_{10}$ $\downarrow$  & FID$_{12}$ $\downarrow$  \\  
\hline

CMR~\cite{kanazawa2018learning} & & -  & 0.703 & 140.9 & 176.2 & 180.1  \\
UMR~\cite{li2020self} & & - & 0.734 & 40.0 & 72.8 & 86.9 \\
U-CMR~\cite{goel2020shape} & & - & 0.701 & 65.0 & 314.9 & 315.2  \\
View-gen~\cite{bhattad2021view} $^{\dagger}$ & & - & 0.629 & - & - & 70.3  \\
SMR~\cite{hu2021self} & & estimated & 0.751 & 55.9 & 65.7 & 85.6  \\
SMR & & ground truth & \textbf{0.800} & 52.9 & 63.2 & 79.3  \\
\hline
CMR  & \checkmark & estimated & 0.717 &  121.6 &  150.5 & 158.4 \\
UMR  & \checkmark & estimated & 0.739 & 38.8 & 78.2 & 91.3 \\
Ours & \checkmark & estimated & 0.752 & \textbf{37.3} & \textbf{38.7} & \textbf{56.8} \\
\hline
\end{tabular}
\label{tab:benchmark}
\end{table}


\begin{figure}[t]
	\centering
	\includegraphics[width=0.9\linewidth]{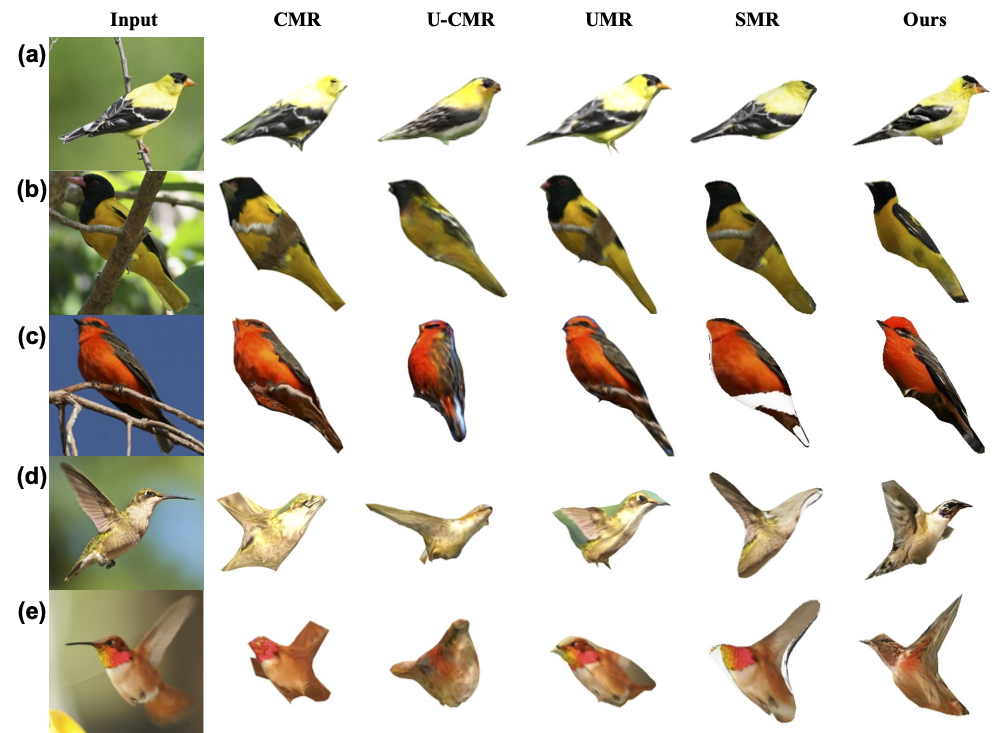}
	\caption{
    Qualitative results on CUB. Our method achieves highly faithful and realistic 3D reconstruction. In particular, it exhibits superior generalization under various challenging scenarios, including with occlusion (b, c) and extended articulation (d, e).
	}
	\label{fig:bird_single}
\end{figure}

\subsection{Comparison with Baselines}
\begin{figure}[t]
	\centering
	\includegraphics[width=1.0\linewidth]{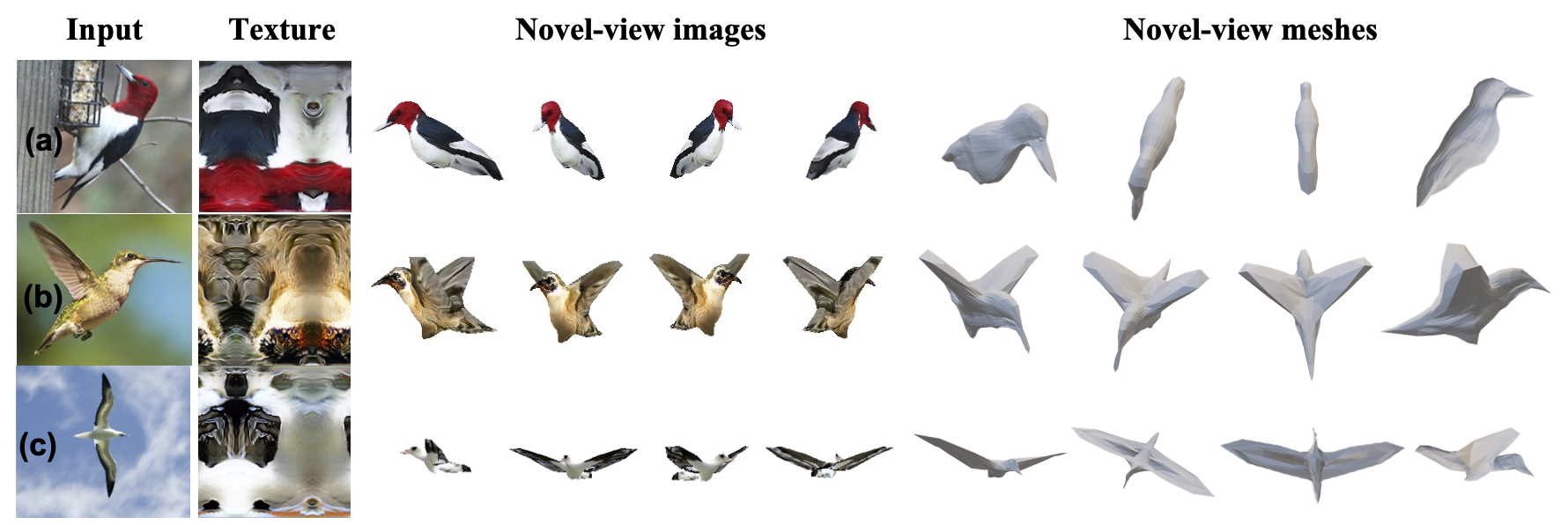}
	\caption{
    Novel-view rendering results on CUB. Our method gives realistic and faithful 3D reconstruction in terms of both 3D shape and appearance. It generalizes fairly well to invisible regions and challenging articulations. }
	\label{fig:bird_multi}
\end{figure}

We compare MeshInversion with various existing methods on the CUB dataset, and report quantitative results in Tab.~\ref{tab:benchmark}. Overall, MeshInversion achieves state-of-the-art results on perceptual metrics, particularly multi-view FIDs, and is on par with existing methods in terms of IoU.
The qualitative results in Fig.~\ref{fig:teaser}, Fig.~\ref{fig:bird_single} and Fig.~\ref{fig:bird_multi} show that MeshInversion achieves highly faithful and realistic 3D reconstruction, particularly when observed from novel views.
Moreover, our method generalizes reasonably well to highly articulated shapes, such as birds with long tails and open wings, where many of the existing methods fail to give satisfactory reconstructions, as Fig.~\ref{fig:teaser} (c)(d) and Fig.~\ref{fig:bird_single} (d)(e) show.
We note that although SMR gives the competitive IoU with estimated mask, it lacks fine geometric details and looks less realistic, \eg, the beak in Fig.~\ref{fig:teaser} (c) and  Fig.~\ref{fig:bird_single}.

\noindent
\textbf{Texture Flow vs. Texture Regression.} Texture flow is extensively adopted in existing methods, except for U-CMR.
Although texture flow-based methods are typically easier to learn and give superior texture reconstruction for visible regions, they tend to give incorrect predictions for invisible regions, \eg, abdomen or back as shown in Fig.~\ref{fig:teaser}. 
In contrast, MeshInversion, which performs direct regression of textures, benefits from a holistic understanding of the objects and gives remarkable performance in the presence of occlusion, while texture flow-based methods only learn to copy from the foreground pixels including the obstacles, \eg, twig, from the bird, as shown in Fig.~\ref{fig:teaser} (a) and Fig.~\ref{fig:bird_single} (b)(c). Due to the same reason, these methods also tend to copy background pixels onto the reconstructed object when the shape prediction is inaccurate, as shown in Fig.~\ref{fig:teaser} (d) and Fig.~\ref{fig:bird_single} (d).
More qualitative results and multi-view comparisons can be found in the supplementary materials.

\noindent
\textbf{Test-time Optimization.} While existing methods mostly adopt an auto-encoder framework and perform inference with a single forward pass, MeshInversion is optimization-based. For a fair comparison, we also introduce test-time optimization (TTO) for baseline methods, if applicable, with access to predicted masks as well. 
Specifically, CMR and UMR have a compact latent code with a dimension of 200, which is desirable for efficient fine-tuning.
As shown in Tab.~\ref{tab:benchmark}, TTO of existing methods overall yields higher fidelity, but our proposed method remains highly competitive in terms of perceptual and geometric performance.
Interestingly, UMR with TTO achieves marginal improvement in terms of IoU and single-view FID at the cost of worsening novel-view FID. 
This further shows the superiority of generative prior captured through adversarial training over that captured in an auto-encoder, including UMR that is coupled with adversarial training, and its effectiveness of such appealing prior in 3D reconstruction.


\setlength{\columnsep}{10pt}
\setlength{\intextsep}{5pt}
\begin{wraptable}{r}{0.55\textwidth}
\centering
\caption{User preference study on CUB in terms of the quality and faithfulness of texture, shape, and overall 3D reconstruction.
}
\small
\begin{tabular}{c|ccccc}
\hline
Criterion & CMR & U-CMR & UMR & SMR & Ours \\
\hline
Texture & 2.7\% & 15.7\% & 13.2\% & 6.6\% & \textbf{61.8\%} \\
Shape & 2.7\% &  19.6\% & 14.6\% & 4.2\% & \textbf{58.9\%} \\
Overall & 2.5\% & 19.8\% & 12.8\% & 3.3\% & \textbf{61.5\%} \\

\hline
\end{tabular}
\label{tab:user_study}
\end{wraptable}

\noindent
\textbf{User Study.} 
We further conduct a user preference study on multi-view renderings of 30 randomly selected birds, and ask 40 users to choose the most realistic and faithful reconstruction in terms of texture, shape, and overall 3D reconstruction.
Tab.~\ref{tab:user_study} shows that MeshInversion gives the the most preferred results, whereas all texture flow-based methods give poor results mainly due to their incorrect prediction for unseen regions.

\begin{figure}[ht] 
	\begin{minipage}{0.69\textwidth}
	\includegraphics[width=0.98\linewidth]{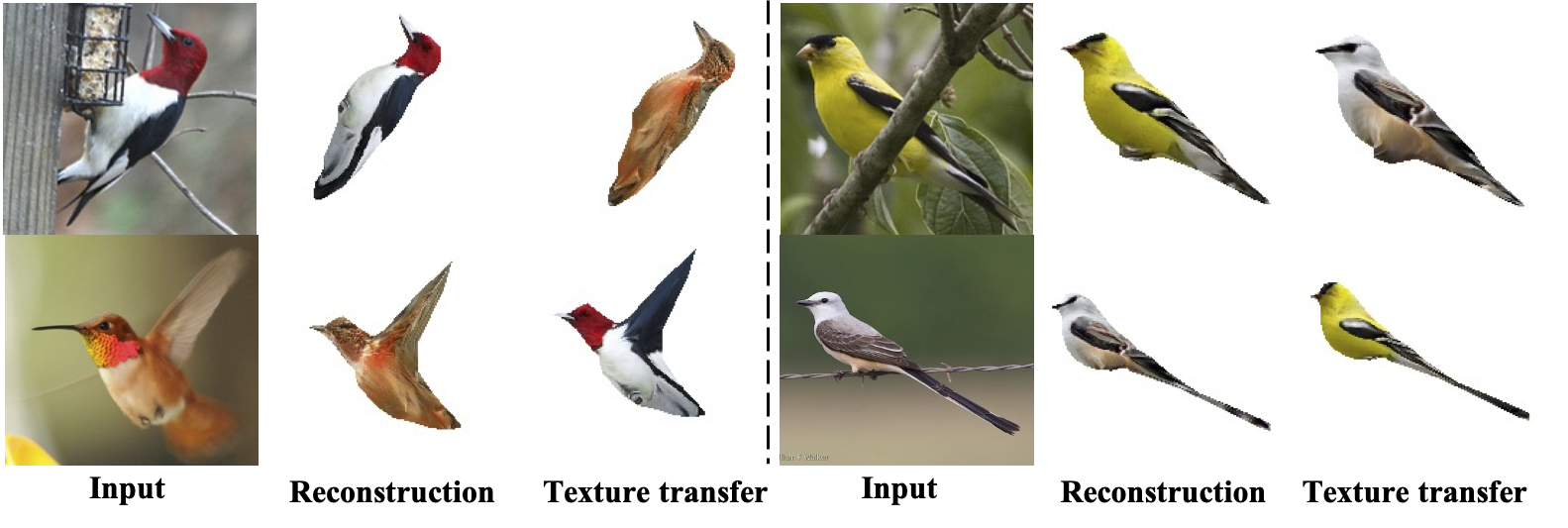}
	\end{minipage} \hspace{0.01\textwidth}
	\begin{minipage}{0.27\textwidth}
		\caption{Our method enables faithful and realistic texture transfer between bird instances even with highly articulated shapes.
	}
    \label{fig:tex_swap}
	\end{minipage} \hspace{0.01\textwidth}
\end{figure}

\subsection{Texture Transfer}
As the shape and texture are topologically and semantically aligned in the UV space, it allows us to modify the surface appearance across bird instances. In Fig.~\ref{fig:tex_swap}, we sample pairs of instances and swap their texture maps. Thanks to the categorical semantic consistency, the resulting new 3D objects remain highly realistic even for extended articulations like open wings and long tails.


\begin{figure}[ht]
	\centering
	\includegraphics[width=0.99\linewidth]{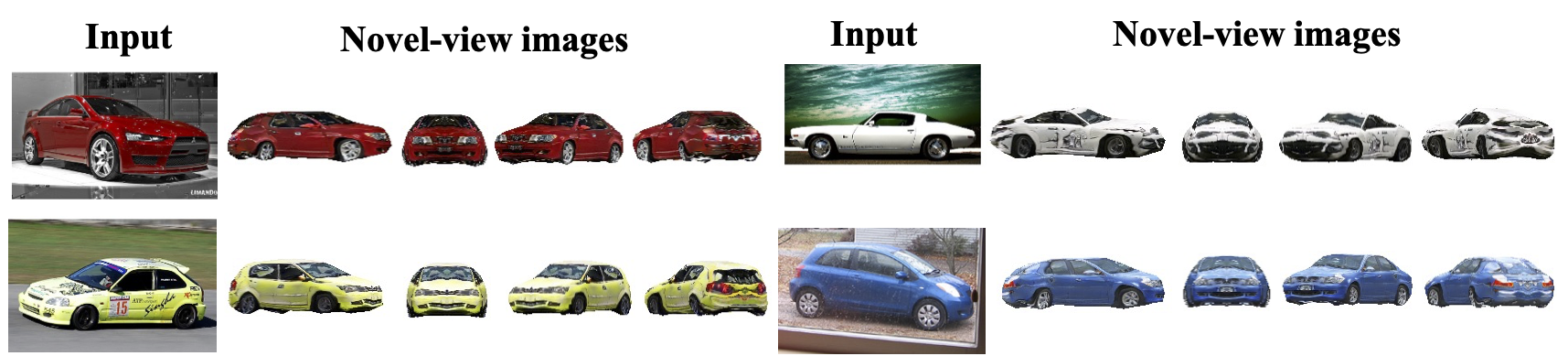}
	\caption{
    Qualitative results on PASCAL3D+ Car. Our method gives reasonably good performance across different car models and appearances.
	}
	\label{fig:car}
\end{figure}

\subsection{Evaluation on PASCAL3D+ Car}

\setlength{\columnsep}{10pt}
\setlength{\intextsep}{5pt}
\begin{wraptable}{r}{0.52\textwidth}
\setlength{\tabcolsep}{4pt}
\centering
\caption{3D IoU on PASCAL3D+ Car. Both the deformable model fitting-based method CSDM~\cite{kar2015category} and volume-based method DRC~\cite{tulsiani2017multi} do not predict object texture.
}
\vspace{8pt}
\begin{tabular}{c|cccc}
\hline
{} & CSDM & DRC & CMR & Ours \\
\hline 
3D IoU & 0.60  & \textbf{0.67} & 0.64 & 0.66 \\ 
\hline
\end{tabular}
\label{tab:3d_iou}
\end{wraptable}
We also evaluate MeshInversion on the man-made rigid car category. As demonstrated in Fig.~\ref{fig:car}, it performs reasonably well across different car models and appearances. Unlike \cite{bhattad2021view} and \cite{goel2020shape} that explicitly use one or multiple mesh templates provided by PASCAL3D+, the appealing GAN prior implicitly provides a rich number of templates that makes it possible to reconstruct cars of various models.
Given the approximated 3D ground truths in PASCAL3D+, we show in Tab~\ref{tab:3d_iou} that MeshInversion performs comparably to baselines in terms of 3D IoU.


\begin{table}
\setlength{\tabcolsep}{4pt}
\centering
\caption{Ablation study. Compared to conventional texture and mask losses, our proposed pixel-level Chamfer texture losses $\mathcal{L}_{pCT}$ and feature-level one $\mathcal{L}_{fCT}$, and Chamfer mask loss $\mathcal{L}_{CM}$ are effective to address the challenges due to misalignment and quantization during rendering. Despite using not-so-accurate camera poses, our method gives compelling geometric and perceptual performance by jointly optimizing 3D shape and camera during GAN inversion.}
\small
\begin{tabular}{ccc|cccc}
\hline
Mask loss & Texture loss & Camera & IoU $\uparrow$ & FID$_{1}$ $\downarrow$  & FID$_{10}$ $\downarrow$ & FID$_{12}$ $\downarrow$   \\  
\hline
IoU loss &   L1 + perceptual loss & fine-tuned & 0.580 & 82.8 & 78.3 & 92.01 \\
\hline
$L_{CM}$ &   L1 loss  & fine-tuned & 0.732 & 58.9 & 78.3 & 74.7 \\
$L_{CM}$ &   L1 + perceptual loss & fine-tuned & 0.741 & 56.3 & 44.0 & 64.9 \\
$L_{CM}$ &  contextual loss & fine-tuned & 0.718 & 69.0 & 57.7 & 75.5 \\
\hline
L1 loss & $L_{pCT}$ + $L_{fCT}$ & fine-tuned & 0.589  & 71.8  & 73.2 & 74.1  \\
IoU loss & $L_{pCT}$ + $L_{fCT}$  & fine-tuned &  0.604  & 72.1  & 71.8 & 70.5 \\
\hline
\textbf{$L_{CM}$} & \textbf{$L_{pCT}$ + $L_{fCT}$}  & fixed & 0.695 &	40.9 &	41.38 &	60.0 \\
\hline
\textbf{$L_{CM}$} & \textbf{$L_{pCT}$ + $L_{fCT}$}  & fine-tuned & 0.752 & \textbf{37.3} & \textbf{38.7} & \textbf{56.8}  \\
\hline
\end{tabular}
\label{tab:ablation}
\end{table}
\subsection{Ablation Study}
\noindent
\textbf{Effectiveness of Chamfer Mask Loss.} 
As compared to Chamfer mask loss in Tab.~\ref{tab:ablation}, both conventional mask losses give significantly worse reconstruction results. As existing mask losses are obtained through rasterization, the induced information loss often makes them less accurate in capturing subtle shape variations, which undermines geometry recovery. A detailed sensitivity study of various mask losses can be found in the supplementary materials.

\noindent
\textbf{Effectiveness of Chamfer Texture Loss.} In the presence of imperfect poses and high-frequency details in textures, we show in Tab.~\ref{tab:ablation} that our proposed pixel- and feature-level Chamfer texture losses are highly effective compared to existing losses. In particular, pixel-to-pixel L1 loss tends to give blurry reconstructions. Feature-based losses, perceptual loss~\cite{johnson2016perceptual} and contextual loss~\cite{mechrez2018contextual}, are generally more robust to misalignment, but they are usually not discriminative enough to reflect the appearance details between two images. Although the contextual loss is designed to address the misalignment issue for image-to-image translation, it only considers feature distances while ignoring their positions in the image.

\noindent
\textbf{Effectiveness of Optimizing Camera Poses.} In Tab.~\ref{tab:ablation}, we show that compared to only optimizing latent code, jointly fine-tuning imperfect camera poses effectively improves geometric performance.

\section{Discussion}
\label{sec:discussion}
We have presented a new approach for monocular 3D object reconstruction. It exploits generative prior encapsulated in a pre-trained GAN and reconstructs textured shapes through GAN inversion. To address reprojection misalignment and discretization-induced information loss due to 3D-to-2D degradation, we propose two Chamfer-based losses in the 2D space, \ie, Chamfer texture loss and Chamfer mask loss. By efficiently incorporating the GAN prior, MeshInversion achieves highly realistic and faithful 3D reconstruction, and exhibits superior generalization power for challenging cases, such as in the presence of occlusion or extended articulations.
However, this challenging problem is far from being solved. In particular, although we can faithfully reconstruct flying birds with open wings, the wings are only represented by a few vertices due to semantic consistency across the entire category, which strictly limits the representation power in terms of geometry and texture details. Therefore, future work may explore more flexible solutions, for instance, an adaptive number of vertices can be assigned to articulated regions to accommodate richer details.

\noindent
\textbf{Acknoledgement}. This study is supported under the RIE2020 Industry Alignment Fund – Industry Collaboration Projects (IAF-ICP) Funding Initiative, Singapore MOE AcRF Tier 2 (MOE-T2EP20221-0011), Shanghai AI Laboratory, as well as cash and in-kind contribution from the industry partners.

\appendix
\renewcommand\thesection{\Alph{section}}
\section{Implementation Details} 
\noindent
\textbf{Architectures.} We follow the same architectures for the generator and the UV space discriminator as described in ConvMesh~\cite{pavllo2020convolutional} for pre-training.
ConvMesh baseline uses a convolutional generator $G$ with two branches, to generate deformation map $\mathbf{S} \in \Rbb^{32\times32}$ and texture map $\mathbf{T} \in \Rbb^{512\times512}$ in the UV space respectively from a latent code $\mathbf{z} \in \Rbb^{64}$. The UV space discriminator consists of two sub-discriminators, for discriminating the deformation map and texture map respectively. In addition, we introduce an image space discriminator to further enforce the realism of the synthesized texture and shape, following the architecture of PatchGAN~\cite{isola2017image}. We render the synthesized textured mesh to images with the DIB-R differentiable renderer~\cite{chen2019learning} following the Kaolin~\cite{Kaolin} implementation.

\noindent
\textbf{Preparation of Pseudo Ground Truths.} During pre-training, discrimination in the UV space requires pseudo ground truth deformation maps and texture maps of the training images. The pseudo deformation maps are obtained by training a mesh reconstruction baseline on the training set. It outputs a 3D object as a deformation map and a texture map as well.
Subsequently, the associated pseudo texture maps are obtained through a form of inverse rendering, \ie, projecting the foreground pixels from the natural images onto the UV space. Since only the visible regions can be projected to the UV space, all the resulting pseudo texture maps are in fact partial. During training, the generated texture maps are masked to form partial texture maps as well prior to discrimination. More details of pseudo ground truths preparation can be found in \cite{pavllo2020convolutional}.

Note that the mesh reconstruction baseline model is overfitted for shape estimation to the training set and is not suitable for 3D reconstruction. In particular, the resulting network generally gives blurry textures, and the shape estimation does not generalize very well to unseen images. Quantitatively, it gives an IoU of 0.671 in contrast to our method with an IoU of 0.752.

\noindent
\textbf{Pre-training.} 
We pre-train ConvMesh following a class conditional setting for 600 epochs, with a batch size of 128, which takes 15 hours on four Nvidia V100 GPUs.
The generator is updated once every three iterations with a learning rate of $1\times 10^{-4}$ whereas the discriminators are updated concurrently twice every three iterations with a learning rate of $4\times 10^{-4}$. We use the Adam optimizer~\cite{kingma2014adam} with $\beta_1=0$ and $\beta_2=0.9$. For the objective function $\mathcal{L}_{G}$ in the main paper, we let $\lambda_{uv}=1$ and $\lambda_{I}=0.04$. We use the same settings for CUB and PASCAL 3D+.

Thanks to the discriminator in the image space, our pre-training results are better than the baseline with a clear margin, as shown in Tab.~\ref{tab:fid_sigma}.

\setlength{\tabcolsep}{6pt}
\begin{table}[h]
\centering
\label{tab:fid_sigma}
\caption{Pre-training results on CUB comparing ConvMesh baseline and ours with the image space discrimination. The Full FID is computed on generated meshes and generated textures; the Texture FID is computed on the generated texture and mesh estimated using the differentiable renderer; the Mesh FID is computed on the pseudo ground truth texture with predicted mesh. We report FID with truncated $\sigma=0.25$. }
\begin{tabular}{c|ccc}
\hline
{} & Full & Texture & Mesh \\
\hline
ConvMesh baseline & 33.6 & 28.7 & 19.5 \\
Ours w/ image space discrimination & \textbf{28.3} & \textbf{27.2} & \textbf{18.7} \\
\hline
\end{tabular}
\end{table}

\noindent
\textbf{Inversion.} We adapt a multi-stage inversion with different learning rates, with learning rates of the latent code $[1\times 10^{-1},5\times 10^{-2},1\times 10^{-2},5\times 10^{-3}]$, learning rates of the camera pose $[1\times 10^{-2},5\times 10^{-3},1\times 10^{-3},5\times 10^{-4}]$, and iterations $[50, 50, 50, 50]$. We use the Adam optimizer with $\beta_1=0$ and $\beta_2=0.99$.
For each testing instance, the inversion process takes around 40 seconds, and our framework supports distributed inference.
For the overall inversion loss $\mathcal{L}_{inv}$, we set the weights of $\mathcal{L}_{pCT}$, $\mathcal{L}_{fCT}$, $\mathcal{L}_{CM}$, $\mathcal{L}_{smooth}$ and $\mathcal{L}_{z}$ as 1, 0.05, 10, 0.00005 and 0.05 respectively. For the Chamfer texture losses, we let $\epsilon_s=0.9$, $\epsilon_a=1$, and $\alpha=1$. In particular, $1-\epsilon_s$ corresponds to degree of tolerance to local misalignment. We show in Tab.~\ref{tab:tex_eps} that a smaller $\epsilon_s$ gives relatively better results in the presence of imperfect camera poses and high-frequency details. One can tighten $\epsilon_s$ if more accurate camera poses are given.

\begin{table}
\setlength{\tabcolsep}{6pt}
\centering
\caption{Effect of $\epsilon_s$ in Chamfer texture losses.}
\small
\begin{tabular}{c|cccc}
\hline
$\epsilon_s$ & IoU $\uparrow$ & FID$_{1}$ $\downarrow$  & FID$_{10}$ $\downarrow$ & FID$_{12}$ $\downarrow$   \\  
\hline
0.999	&				0.747 &	38.7 &	39.8  &	58.0 \\
0.99	&				0.746 &	38.6 &	40.0 &	58.3 \\
0.98	&				0.748 &	38.7 &	39.7 &	57.8 \\
0.95	&				0.749 &	37.6 &	39.0 &	57.2 \\
0.9		&			\textbf{0.752} &	\textbf{37.3} &	\textbf{38.7} &	\textbf{56.8} \\
\hline
\end{tabular}
\label{tab:tex_eps}
\end{table}

\noindent
\textbf{Test-time Optimization of Baselines.}
Similar to GAN inversion, a relatively compact latent space is desirable for efficient optimization during the test time. Both CMR and UMR have a latent code with a dimension of 200. In contrast, U-CMR has a latent code with a dimension of 4096, whereas SMR does not follow an auto-encoder architecture, but directly encodes the 3D attributes from the image with the associated mask. Therefore, SMR and U-CMR are infeasible to be adapted for test-time optimization.

We adapt CMR and UMR as follows during the test time: 
The latent code is first obtained with a single forward pass by the image encoder, and the camera pose is then obtained by the camera pose estimator.
We then fine-tune the latent code and the camera pose by minimizing the mask loss and texture loss by comparing against the estimated mask and the input image respectively, where the network weights remain fixed. The choices of loss functions and their weights follow those during the training time. For an equal comparison, we fine-tune with the same Adam optimizer and for the same number of iterations, 200, for each testing instance. Since our method uses randomly initialized latent code whereas the forward pass by the image encoder already provides a good initialization, we use a smaller learning rate for the latent code, $5\times 10^{-3}$. For the camera pose, we use the same learning rates as our method following the multi-stage setting.

\noindent
\textbf{Additional Details for User Study.}
We conduct a user preference study on CUB to evaluate our method. This user preference study involves 40 users, 30 objects, and five methods (four baselines and ours).
The 40 users are invited from several different backgrounds, including finance, business, life science, and information technology. 
We randomly choose 30 objects from the testing split, and ensure the following varieties are contained in the selection: complex and a wide range of texture, with highly articulated shapes, and in the presence of occlusion, etc.
The reconstructed 3D objects are rendered from three different viewpoints to make sure that the entire object is observable by the user.
For each input image, we give users unlimited time to select the method that gives the most faithful and realistic result in terms of three separate criteria: texture quality, shape quality, and overall textured shape reconstruction. 

\section{Comparison with Shelf-Sup}
Unlike mainstream approaches compared in the main paper that directly deform a category-level template mesh, Shelf-Sup~\cite{ye2021shelf} first gives a coarse volumetric prediction, and then converts the coarse volume into a mesh followed by test-time optimization. This design demonstrates its scalability on significantly more categories with various topologies (though still category-specific). However,
such scalability is at the sacrifice of categorical
semantic consistency without a mesh template, leading to less
compelling reconstructions on birds, as compared in Tab.~\ref{tab:shelf-sup}. Both Shelf-Sup and UMR employ adversarial loss when training the auto-encoders for reconstruction, whereas the generative adversarial training in GANs for 3D object synthesis offers richer prior, giving better generalization and photo-realism as demonstrated in Sec 4.1. In addition, Shelf-Sup also tends to suffer from blurry reconstructions with imperfect camera predictions, which calls for the robustness offered by Chamfer texture loss.

\setlength{\tabcolsep}{4.0 pt}
\begin{table}[h]
\caption{Comparison with Shelf-Sup on CUB.}
\begin{center}
\begin{tabular}{l|c|c|c|c}
\hline
Methods & IoU $\uparrow$ & FID$_{1}$ $\downarrow$  & FID$_{10}$ $\downarrow$  & FID$_{12}$ $\downarrow$ \\
\hline
Shel-Sup & 0.707 & 81.2 & 140.1 & 161.0 \\
Ours & \textbf{0.752} & \textbf{37.3} & \textbf{38.7} & \textbf{56.8} \\
\hline
\end{tabular}
\end{center}
\label{tab:shelf-sup}
\end{table}


\setlength{\columnsep}{10pt}
\setlength{\intextsep}{5pt}
\begin{figure}
  \centering
  \includegraphics[width=.90\textwidth]{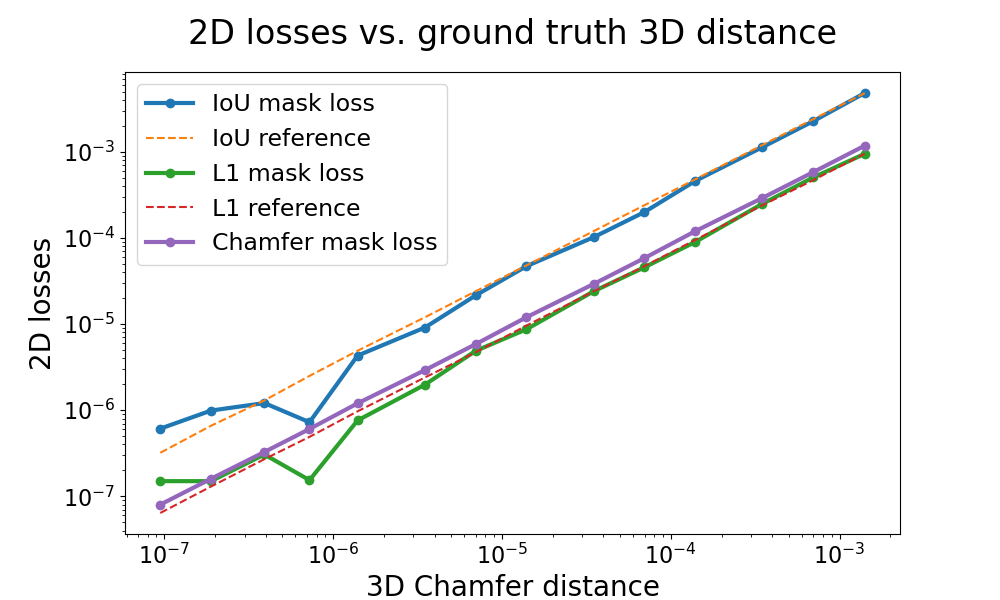}
  \caption{\small Sensitivity comparison of various mask losses. Unlike existing mask losses, the constant slope by Chamfer mask loss implies its high accuracy across a wide range of shape variation. 
  Each data point is the mean of 100 simulation runs.
  }
  \label{fig:mask_slope}
\end{figure}

\section{Sensitivity Comparison of Mask Losses}
We quantitatively analyze the sensitivity of various mask losses by measuring the distance between two 3D shapes at different degrees of shape variations.
Specifically, we utilize the pre-trained ConvMesh to randomly generate 100 3D shapes. For each shape $O_i$, we introduce a variation of the shape by jittering its latent code $z_i$ by a step size $\eta$ at a random direction, giving the deviated shape $O_i^{\prime}$. 
We then measure the ground truth distance in the 3D space between $O_i$ and $O_i^{\prime}$ using Chamfer distance, and compute the distances in the 2D space using IoU loss, L1 loss, and Chamfer mask loss respectively. Specifically, we vary the step size $\eta$ from $1\times 10^{-6}$ all the way to $1\times 10^{-1}$, giving 3D Chamfer distances from $1\times 10^{-7}$ to $1\times 10^{-3}$. Given these generated and jittered 3D shape pairs, we plot various 2D mask losses against ground truth 3D Chamfer distance in Fig.~\ref{fig:mask_slope}.
As all the four metrics, including Chamfer distances, are L1-like, all the three 2D losses should be linearly correlated to the 3D Chamfer distance, \ie, these plots should maintain constant slopes. However, due to discretization-induced information loss, both IoU mask loss and L1 mask loss are not able to accurately reflect the subtle variation in the 3D shape.  Such inaccuracy eventually leads to insensitive gradients and undermines geometry recovery. In contrast, $\mathcal{L}_{CM}$ intercepts the rasterization process and naturally retains information.

\section{Qualitative Results for Texture Loss Ablation}
\begin{figure}[ht]
\centering \includegraphics[width=0.90\linewidth]{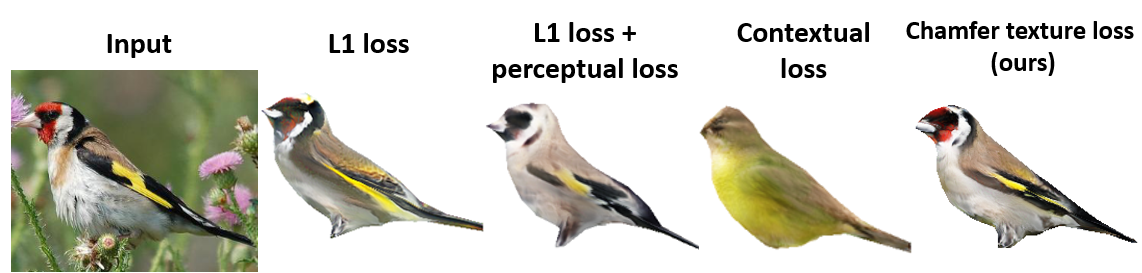}
\caption{
     Ablation study on Chamfer texture losses.
	}
\label{fig:texture_qual_results}
\end{figure}

We provide qualitative results for ablation study on texture losses in Fig.~\ref{fig:texture_qual_results} (zoom in for details). 
As inaccurate camera initialization easily leads to reprojection misalignment, in the presence of high-frequency texture, conventional pixel-aligned appearance losses, \eg, L1 loss on RGB images and perceptual loss on feature maps, would tend to give noisy supervision signal. This often leads to blurry reconstructions, especially for L1 loss. 
While earlier attempt to address the misalignment issue by contextual loss treats feature maps as sets of feature vectors,
it totally ignores the feature locations in the image coordinate, resulting in unfaithful reconstructions.

\section{Additional Qualitative Results}
\begin{figure}[ht]
	\centering
	\includegraphics[width=1.0\linewidth]{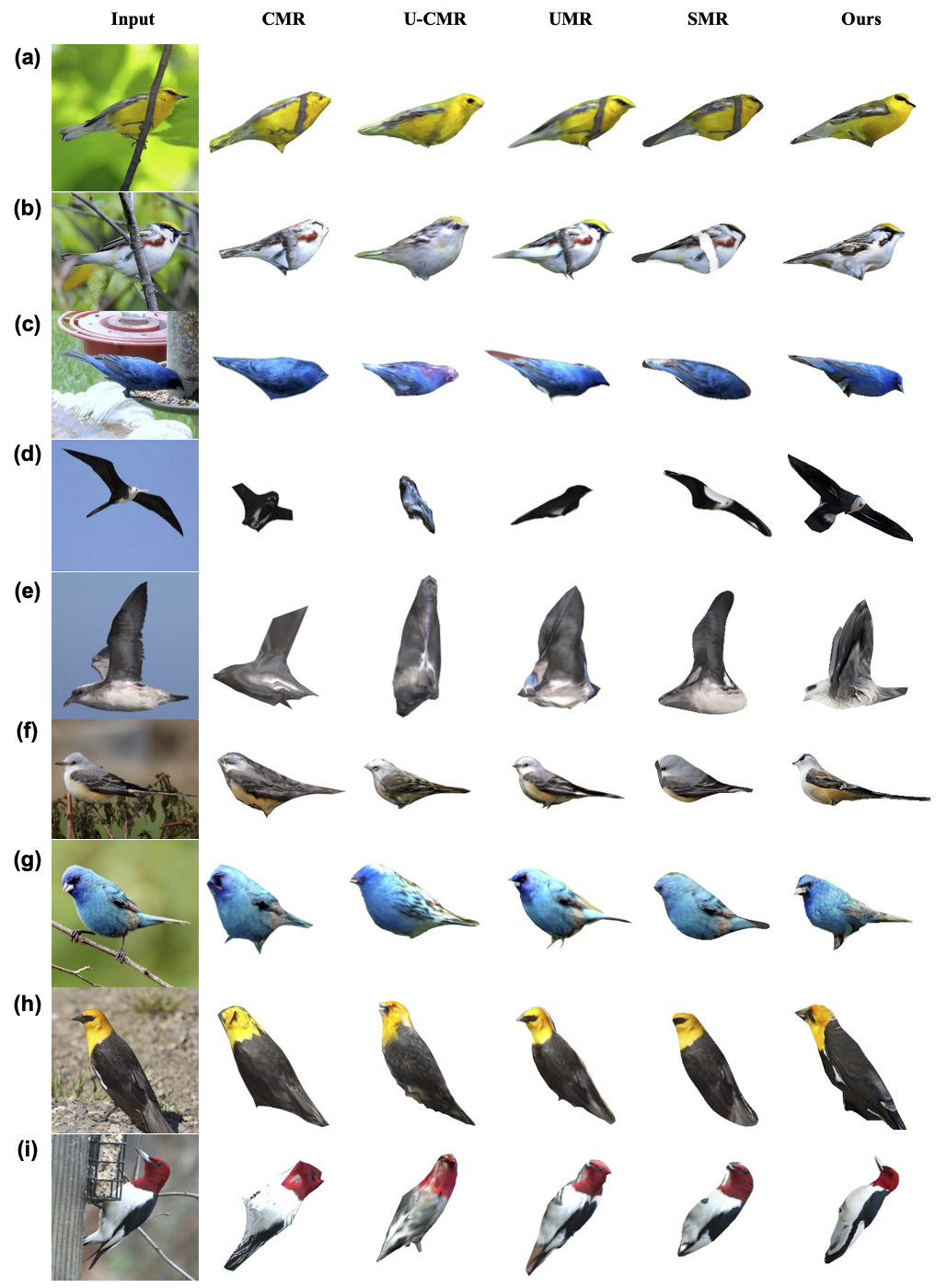}
	\caption{
     Additional qualitative results on CUB.
	}
	\label{fig:bird_2}
\end{figure}

\begin{figure}[ht]
	\centering
	\includegraphics[width=1.0\linewidth]{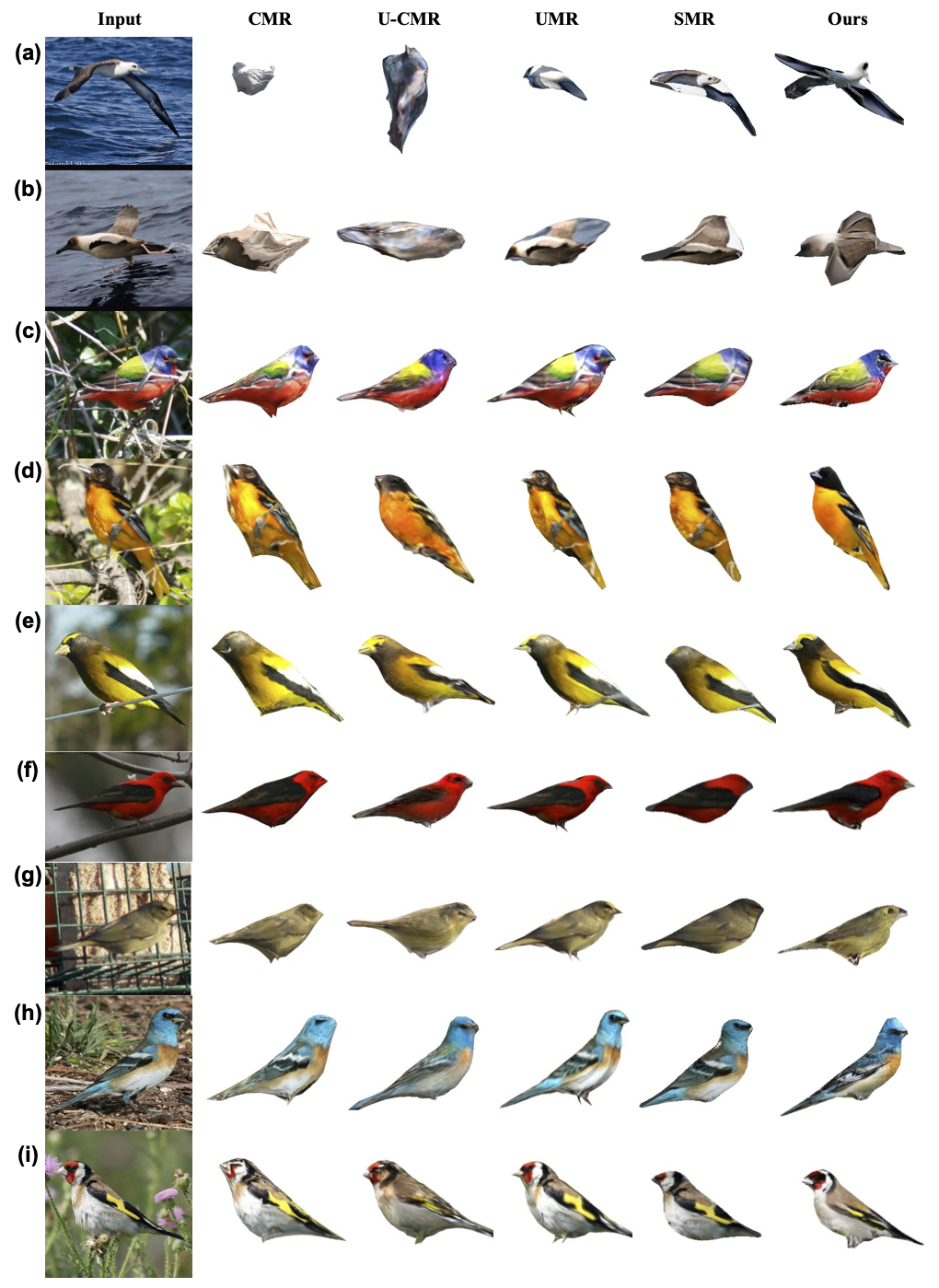}
	\caption{
     Additional qualitative results on CUB (continued).
	}
	\label{fig:bird_3}
\end{figure}

We provide more single-view illustrative examples for CUB in Fig.~\ref{fig:bird_2} and Fig.~\ref{fig:bird_3}. In addition, we further demonstrate the merit of our method through 360-degree comparisons with baselines in the supplemental video. The supplemental video also includes illustration of the inversion process, and 360-degree results for texture transfer and on PASCAL3D+ Car.

\clearpage
%
%
\bibliographystyle{splncs04}
\bibliography{egbib}

\begin{thebibliography}{10}
\providecommand{\url}[1]{\texttt{#1}}
\providecommand{\urlprefix}{URL }
\providecommand{\doi}[1]{https://doi.org/#1}

\bibitem{achlioptas2018learning}
Achlioptas, P., Diamanti, O., Mitliagkas, I., Guibas, L.: Learning
  representations and generative models for {3D} point clouds. In: ICML (2018)

\bibitem{bau2020semantic}
Bau, D., Strobelt, H., Peebles, W., Zhou, B., Zhu, J.Y., Torralba, A., et~al.:
  Semantic photo manipulation with a generative image prior. In: SIGGRAPH
  (2019)

\bibitem{bau2019seeing}
Bau, D., Zhu, J.Y., Wulff, J., Peebles, W., Strobelt, H., Zhou, B., Torralba,
  A.: Seeing what a {GAN} cannot generate. In: ICCV (2019)

\bibitem{bhattad2021view}
Bhattad, A., Dundar, A., Liu, G., Tao, A., Catanzaro, B.: View generalization
  for single image textured {3D} models. In: CVPR (2021)

\bibitem{brock2018large}
Brock, A., Donahue, J., Simonyan, K.: Large scale {GAN} training for high
  fidelity natural image synthesis. In: ICLR (2019)

\bibitem{chen2019learning}
Chen, W., Ling, H., Gao, J., Smith, E., Lehtinen, J., Jacobson, A., Fidler, S.:
  Learning to predict {3D} objects with an interpolation-based differentiable
  renderer. In: NeurIPS (2019)

\bibitem{ye2021shelf}
\etal, Y.: Shelf-supervised mesh prediction in the wild. In: CVPR (2021)

\bibitem{gecer2019ganfit}
Gecer, B., Ploumpis, S., Kotsia, I., Zafeiriou, S.: {GANFIT}: Generative
  adversarial network fitting for high fidelity {3D} face reconstruction. In:
  CVPR (2019)

\bibitem{girdhar2016learning}
Girdhar, R., Fouhey, D.F., Rodriguez, M., Gupta, A.: Learning a predictable and
  generative vector representation for objects. In: ECCV (2016)

\bibitem{goel2020shape}
Goel, S., Kanazawa, A., Malik, J.: Shape and viewpoint without keypoints. In:
  ECCV (2020)

\bibitem{goodfellow2014generative}
Goodfellow, I., Pouget-Abadie, J., Mirza, M., Xu, B., Warde-Farley, D., Ozair,
  S., Courville, A., Bengio, Y.: Generative adversarial nets. In: NeurIPS
  (2014)

\bibitem{gu2020image}
Gu, J., Shen, Y., Zhou, B.: Image processing using multi-code {GAN} prior. In:
  CVPR (2020)

\bibitem{henderson2020leveraging}
Henderson, P., Tsiminaki, V., Lampert, C.H.: Leveraging {2D} data to learn
  textured {3D} mesh generation. In: CVPR (2020)

\bibitem{heusel2017gans}
Heusel, M., Ramsauer, H., Unterthiner, T., Nessler, B., Hochreiter, S.: Gans
  trained by a two time-scale update rule converge to a local nash equilibrium.
  In: NeurIPS (2017)

\bibitem{hu2021self}
Hu, T., Wang, L., Xu, X., Liu, S., Jia, J.: Self-supervised {3D} mesh
  reconstruction from single images. In: CVPR (2021)

\bibitem{isola2017image}
Isola, P., Zhu, J.Y., Zhou, T., Efros, A.A.: Image-to-image translation with
  conditional adversarial networks. In: CVPR (2017)

\bibitem{Kaolin}
Jatavallabhula, K.M., Smith, E., Lafleche, J.F., Tsang, C.F., Rozantsev, A.,
  Chen, W., Xiang, T., Lebaredian, R., Fidler, S.: Kaolin: A {PyTorch} library
  for accelerating {3D} deep learning research. CoRR  \textbf{abs/1911.05063}
  (2019)

\bibitem{johnson2016perceptual}
Johnson, J., Alahi, A., Fei-Fei, L.: Perceptual losses for real-time style
  transfer and super-resolution. In: ECCV (2016)

\bibitem{kanazawa2018end}
Kanazawa, A., Black, M.J., Jacobs, D.W., Malik, J.: End-to-end recovery of
  human shape and pose. In: CVPR (2018)

\bibitem{kanazawa2018learning}
Kanazawa, A., Tulsiani, S., Efros, A.A., Malik, J.: Learning category-specific
  mesh reconstruction from image collections. In: ECCV (2018)

\bibitem{kar2015category}
Kar, A., Tulsiani, S., Carreira, J., Malik, J.: Category-specific object
  reconstruction from a single image. In: CVPR (2015)

\bibitem{karras2019style}
Karras, T., Laine, S., Aila, T.: A style-based generator architecture for
  generative adversarial networks. In: CVPR (2019)

\bibitem{kingma2014adam}
Kingma, D.P., Ba, J.: Adam: A method for stochastic optimization. In: ICLR
  (2015)

\bibitem{kingma2013auto}
Kingma, D.P., Welling, M.: Auto-encoding variational bayes. In: ICLR (2014)

\bibitem{kirillov2020pointrend}
Kirillov, A., Wu, Y., He, K., Girshick, R.: {PointRend}: Image segmentation as
  rendering. In: CVPR (2020)

\bibitem{li2020self}
Li, X., Liu, S., Kim, K., De~Mello, S., Jampani, V., Yang, M.H., Kautz, J.:
  Self-supervised single-view {3D} reconstruction via semantic consistency. In:
  ECCV (2020)

\bibitem{lin2017focal}
Lin, T.Y., Goyal, P., Girshick, R., He, K., Doll{\'a}r, P.: Focal loss for
  dense object detection. In: ICCV (2017)

\bibitem{lin2014microsoft}
Lin, T.Y., Maire, M., Belongie, S., Hays, J., Perona, P., Ramanan, D.,
  Doll{\'a}r, P., Zitnick, C.L.: Microsoft {COCO}: Common objects in context.
  In: ECCV (2014)

\bibitem{lipton2017precise}
Lipton, Z.C., Tripathi, S.: Precise recovery of latent vectors from generative
  adversarial networks. CoRR  \textbf{arXiv:1702.04782} (2017)

\bibitem{liu2019soft}
Liu, S., Chen, W., Li, T., Li, H.: Soft rasterizer: Differentiable rendering
  for unsupervised single-view mesh reconstruction. In: ICCV (2019)

\bibitem{ma2018invertibility}
Ma, F., Ayaz, U., Karaman, S.: Invertibility of convolutional generative
  networks from partial measurements. In: NeurIPS (2018)

\bibitem{mao2017least}
Mao, X., Li, Q., Xie, H., Lau, R.Y., Wang, Z., Paul~Smolley, S.: Least squares
  generative adversarial networks. In: ICCV (2017)

\bibitem{mechrez2018contextual}
Mechrez, R., Talmi, I., Zelnik-Manor, L.: The contextual loss for image
  transformation with non-aligned data. In: ECCV (2018)

\bibitem{mescheder2019occupancy}
Mescheder, L., Oechsle, M., Niemeyer, M., Nowozin, S., Geiger, A.: Occupancy
  networks: Learning {3D} reconstruction in function space. In: CVPR (2019)

\bibitem{niemeyer2020differentiable}
Niemeyer, M., Mescheder, L., Oechsle, M., Geiger, A.: Differentiable volumetric
  rendering: Learning implicit {3D} representations without {3D} supervision.
  In: CVPR (2020)

\bibitem{oechsle2021unisurf}
Oechsle, M., Peng, S., Geiger, A.: {UNISURF}: Unifying neural implicit surfaces
  and radiance fields for multi-view reconstruction. In: ICCV (2021)

\bibitem{pan2019deep}
Pan, J., Han, X., Chen, W., Tang, J., Jia, K.: Deep mesh reconstruction from
  single rgb images via topology modification networks. In: ICCV (2019)

\bibitem{pan20202d}
Pan, X., Dai, B., Liu, Z., Loy, C.C., Luo, P.: Do {2D} gans know {3D} shape?
  unsupervised {3D} shape reconstruction from {2D} image {GANs}. In: ICLR
  (2021)

\bibitem{pan2021exploiting}
Pan, X., Zhan, X., Dai, B., Lin, D., Loy, C.C., Luo, P.: Exploiting deep
  generative prior for versatile image restoration and manipulation. PAMI
  (2021)

\bibitem{pavllo2020convolutional}
Pavllo, D., Spinks, G., Hofmann, T., Moens, M.F., Lucchi, A.: Convolutional
  generation of textured {3D} meshes. In: NeurIPS (2020)

\bibitem{rematas2021sharf}
Rematas, K., Martin-Brualla, R., Ferrari, V.: {ShaRF}: Shape-conditioned
  radiance fields from a single view. In: ICML (2021)

\bibitem{sanyal2019learning}
Sanyal, S., Bolkart, T., Feng, H., Black, M.J.: Learning to regress {3D} face
  shape and expression from an image without {3D} supervision. In: CVPR (2019)

\bibitem{shu20193d}
Shu, D.W., Park, S.W., Kwon, J.: {3D} point cloud generative adversarial
  network based on tree structured graph convolutions. In: ICCV (2019)

\bibitem{simonyan2014very}
Simonyan, K., Zisserman, A.: Very deep convolutional networks for large-scale
  image recognition. In: ICLR (2015)

\bibitem{smith2017improved}
Smith, E.J., Meger, D.: Improved adversarial systems for {3D} object generation
  and reconstruction. In: CoRL (2017)

\bibitem{tulsiani2017multi}
Tulsiani, S., Zhou, T., Efros, A.A., Malik, J.: Multi-view supervision for
  single-view reconstruction via differentiable ray consistency. In: CVPR
  (2017)

\bibitem{wah2011caltech}
Wah, C., Branson, S., Welinder, P., Perona, P., Belongie, S.: The caltech-ucsd
  birds-200-2011 dataset  (2011)

\bibitem{wang2018pixel2mesh}
Wang, N., Zhang, Y., Li, Z., Fu, Y., Liu, W., Jiang, Y.G.: {Pixel2Mesh}:
  Generating {3D} mesh models from single rgb images. In: ECCV (2018)

\bibitem{wang2021neus}
Wang, P., Liu, L., Liu, Y., Theobalt, C., Komura, T., Wang, W.: {NeuS}:
  Learning neural implicit surfaces by volume rendering for multi-view
  reconstruction. In: NeurIPS (2021)

\bibitem{wu2016learning}
Wu, J., Zhang, C., Xue, T., Freeman, B., Tenenbaum, J.: Learning a
  probabilistic latent space of object shapes via {3D} generative-adversarial
  modeling. In: NeurIPS (2016)

\bibitem{xiang2014beyond}
Xiang, Y., Mottaghi, R., Savarese, S.: Beyond {PASCAL}: A benchmark for {3D}
  object detection in the wild. In: WACV (2014)

\bibitem{xie2018learning}
Xie, J., Zheng, Z., Gao, R., Wang, W., Zhu, S.C., Wu, Y.N.: Learning descriptor
  networks for {3D} shape synthesis and analysis. In: CVPR (2018)

\bibitem{yariv2020multiview}
Yariv, L., Kasten, Y., Moran, D., Galun, M., Atzmon, M., Basri, R., Lipman, Y.:
  Multiview neural surface reconstruction by disentangling geometry and
  appearance. In: NeurIPS (2020)

\bibitem{zhang2021unsupervised}
Zhang, J., Chen, X., Cai, Z., Pan, L., Zhao, H., Yi, S., Yeo, C.K., Dai, B.,
  Loy, C.C.: Unsupervised {3D} shape completion through {GAN} inversion. In:
  CVPR (2021)

\bibitem{zhu2020domain}
Zhu, J., Shen, Y., Zhao, D., Zhou, B.: In-domain {GAN} inversion for real image
  editing. In: ECCV (2020)

\bibitem{zhu2018visual}
Zhu, J.Y., Zhang, Z., Zhang, C., Wu, J., Torralba, A., Tenenbaum, J., Freeman,
  B.: Visual object networks: Image generation with disentangled {3D}
  representations. In: NeurIPS (2018)

\end{thebibliography}
\end{document}